\documentclass[runningheads]{llncs}
\usepackage{graphicx}
\usepackage{cite}
\usepackage{amsmath,amssymb,amsfonts}
\usepackage{algorithmic}
\usepackage{textcomp}
\usepackage{csvsimple}
\usepackage{booktabs}
\usepackage{xcolor}
\usepackage[T1]{fontenc}

\usepackage{lscape}

\newcommand{\refequation}[1]{Equation~\eqref{#1}}
\newcommand{\reffigure}[1]{\figurename~\ref{#1}}

\newcommand{\reftable}[1]{Table~\ref{#1}}

\begin{document}
\title{Measuring Board Game Distance}
%
%
\author{Matthew Stephenson \and
Dennis J. N. J. Soemers \and
{\'E}ric Piette \and
Cameron Browne}
\authorrunning{M. Stephenson et al.}
%
\institute{Department of Advanced Computing Sciences, Maastricht University, \newline Paul-Henri Spaaklaan 1, 6229 EN, Maastricht, the Netherlands
\email{\{matthew.stephenson,dennis.soemers,eric.piette,cameron.browne\}
@maastrichtuniversity.nl}}
\maketitle              
\begin{abstract}
This paper presents a general approach for measuring distances between board games within the Ludii general game system. These distances are calculated using a previously published set of general board game concepts, each of which represents a common game idea or shared property. Our results compare and contrast two different measures of distance, highlighting the subjective nature of such metrics and discussing the different ways that they can be interpreted.

\end{abstract}

\section{Introduction}

Ludii is a relatively recent general game system that contains a large variety of different board games\cite{Piette_2020_Ludii}. This includes games with stochasticity and hidden information, alternating and simultaneous move formats, between one and sixteen players, piece stacking, team-based scoring, among many other features. Games in Ludii are described using ludemes, which are specific keywords that are defined within the Ludii Game Description Language (L-GDL). While individually simple, these ludemes can be combined to express complex game rules and mechanics. A previous study demonstrated that it is possible to use a game's ludemes to accurately predict the performance of various game-playing heuristics\cite{Stephenson_2021_General}. However, representing a game solely as the set of ludemes within its description can lead to issues.

Because these ludemes are often combined to express more complex rules and mechanics, their specific order and arrangement can dramatically alter a game's behaviour. Just looking at the ludemes that are present within a game's description is often not enough to understand their wider context and intended effect. For example, knowing that a game contains the move ludeme "hop" does not tell us whether this type of move can be done over friendly or enemy pieces. 
We are also not able to detect how frequently "hop" moves occur in typical play, compared to other types of moves. 
To address these and other similar limitations, a set of general board game concepts was proposed \cite{Piette_2021_Concepts}. These concepts were created as a way to identify and extract higher-level features within each game, thus providing a more complete representation.

In this paper, we explore how these concepts can be used to calculate a measure of distance between any two games in Ludii. In addition to providing insight into the types of games currently available within Ludii, being able to measure the distance between two games has a variety of practical applications. One example is the ability to improve the performance of general game playing agents on unknown games, by identifying similar known games with pre-existing knowledge and results. This application has already motivated prior investigations into measuring game distance within other general game systems, including both the Stanford GGP framework \cite{Jung_2021_DistanceBased,Michulke_2012_Distance} and the General Video Game AI framework \cite{Bontrager_2016_Matching,Mendes_2016_Hyperheuristic,Horn_2016_MCTSEA}. Along with this, measures of game distance can be used by recommender systems to suggest new games to users based on their prior preferences and ratings \cite{Kim_2020_Sequential,Zalewski_2019_Recommender}, for transfer learning between similar games \cite{Soemers_2021_Transfer}, to examine the variety of games within a specific subset \cite{Stephenson_2020_Continuous}, or for game reconstruction purposes \cite{Browne_2018_Modern}.

The remainder of this paper is structured as follows. Section \ref{Datasets} describes the games and concepts that will be used. Section \ref{Data Visualisation} provides visualisations of the overall distribution of concept values across all games within Ludii. Section \ref{Game Distance} presents several different approaches for calculating distances between two games using their concept values, and provides two specific examples based on Cosine Similarity and Euclidean distance. Section \ref{Results} summarises and discusses the results of these two distance measures when applied to all pairs of games. Section \ref{Conclusion} summarises our findings and suggests possibilities for future work.

\section{Datasets}
\label{Datasets}

This section describes the two datasets that were used for this study, that of the games within Ludii and their associated concept values. These datasets were obtained from v1.3.2 of the Ludii database, which is publicly available online.\footnote{www.ludii.games/downloads/database-1.3.2.zip}

\subsection{Games}

As of the time of writing, Ludii version 1.3.2 includes 1059 fully playable games. While some of these games also contain multiple options and rulesets for providing different variations of the same base rules, for the sake of simplicity we will only be considering the default version for each game as provided by Ludii. 

Due to the fact that Ludii was developed as part of the Digital Ludeme Project \cite{Browne_2018_Modern}, the majority of the board games it contains are traditional games that date back many hundreds of years. 
Even though a large assortment of modern abstract games have also been implemented within Ludii, this set of games is unlikely to be fully representative of the complete population of different games that exist within the modern board game industry. For example, Ludii does not currently include any card games, even though many modern board games often use cards in some capacity. Nevertheless, Ludii still contains a substantial variety of different abstract games, and an analysis of its full game library is worth performing.


\subsection{Concepts}
Each concept represents a specific property of a game as a single numerical value. 
These concepts can be binary (e.g. if the game contains hidden information), discrete (e.g. the number of players), or continuous (e.g. the likelihood of a game ending in a draw).
The Ludii database currently lists 499 distinct concepts with computed values for every game.
Each of these concepts is associated with one of six categories based on what aspect of the game they represent, see \reftable{categories_table}.

\begin{table}
\caption{Concept Categories}
\vspace{-4mm}
\label{Table:Categories}
\begin{center}
\begin{tabular}{p{3cm} p{6.5cm} r}
\toprule
 Category & Examples & Count \\ 
  \midrule
 Properties & Num Players, Stochastic, Asymmetric & 21 \\  
 Equipment & Mancala Board, Hex Tiling, Dice, Hand & 74 \\  
 Rules & Hop Capture, Turn Ko, Draw Frequency & 302 \\  
 Math & Multiplication, Intersection, Union & 33 \\
 Visual & Go Style, Chess Component, Stack Type & 42 \\  
 Implementation & Playouts Per Second, Moves Per Second  & 27 \\
 \bottomrule
\end{tabular}
\label{categories_table}
\end{center}
\vspace{-2mm}
\end{table}

Concepts also differ in the way they are computed. 
Compilation concepts can be calculated from just the game's description, and will be the same every time they are computed. 
Playout concepts instead require one or more game traces in order to compute them, and will often vary for the same game if different traces are used (although using a large number of traces can reduce this variance). 
For this study, 87 of the concepts used were playout concepts. These were computed for each game using 100 game traces generated from random play, with a maximum limit of 2500 moves per game trace (after which the result is a draw). 

Due to the different value ranges that each concept can take, we decided to normalise each concept to the same scale.
However, one issue with these concepts is that they are susceptible to having outlier values. 
For example, games played on implied "boards" of potentially unbounded size (e.g. Dominoes) are modelled in Ludii using extremely large static boards.
Another example would be the game \emph{Hermit} which, due to the unique way in which each player's score is represented, produces an average score variance of over 100 million points. Due to cases like these, directly applying Min-Max scaling to our concept values would overemphasise these outliers and make all other values irrelevant. To mitigate this problem, we first applied a bi-symmetric log transformation on all concept values \cite{Webber_2012_Bisymmetric}, see \refequation{Eq:log_transform}.

\begin{equation} \label{Eq:log_transform}
f(x) = \left\{
        \begin{array}{ll}
            log_{2}(x+1) & \quad x \geq 0 \\
            -log_{2}(1-x) & \quad x < 0
        \end{array}
    \right.
\end{equation}

This transformation reduces the impact of the more extreme positive and negative values, while ensuring that binary concepts are unaffected. These new transformed values are then normalised using Min-Max scaling, to give the final concept values for each game.
Thus our full dataset consists of 1059$\times$499 matrix, detailing the concept values for each game. 

\section{Data Visualisation}
\label{Data Visualisation}

\begin{figure}
\centerline{\includegraphics[width=0.8\linewidth]{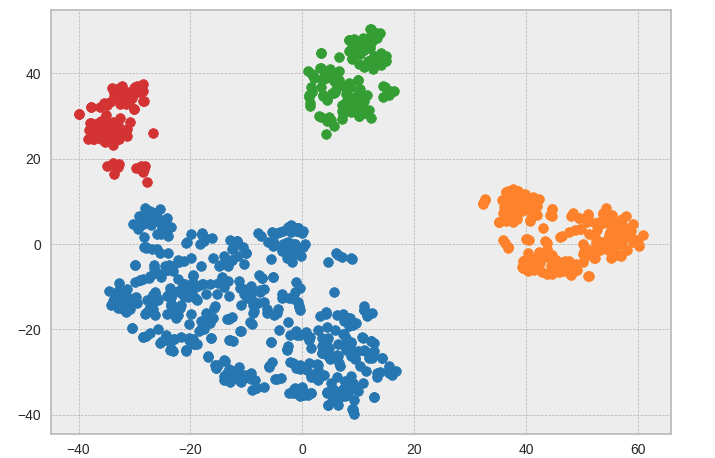}}
\caption{Game concept dataset reduced to two dimensions using t-SNE. Points are coloured based on identified game clusters.}
\label{gameClusters}
\end{figure}

Before calculating any game distances, we decided to first visualise the overall distribution of concept values across all games within Ludii.
To do this, we applied t-distributed stochastic neighbor embedding (t-SNE) \cite{Maaten_2008_TSNE} to reduce our concept dataset to two dimensions, see Figure \ref{gameClusters}.
From this visualisation, we identified four distinct clusters of games. The orange cluster contains 207 games, the green cluster contains 147 games, the red cluster contains 105 games, and the blue cluster contains 600 games.

Analysing these clusters closer reveals some general trends within each group.
\begin{itemize}
    \item The orange cluster contains all games with a Mancala Board, such as \emph{Oware} or \emph{Kalah}. These games are highly separated with a very distinct way of playing, leading to a lot of concepts that are unique to them such as the mechanic of sowing stones. It therefore makes sense that these games would form their own distinct cluster.
    \item The green cluster contains all games with both dice and a track for pieces to move along (e.g. \emph{Backgammon} or \emph{Snakes and Ladders}), as well as a few other games such as \emph{EinStein Würfelt Nicht} and \emph{So Long Sucker}. Despite the clear separation of the games in this cluster from the rest, this distinction is not the sole result of any one game element. Instead, it seems that such games would be better characterised as those with a high degree of uncertainty, either because stochastic elements have a large impact on the game or because of other player's decisions.
    \item The red cluster contains all games with a "Threat" mechanism, predominantly used in Chess-like games when seeing if the king is in Check, as well as some other similar games such as \emph{Ploy} and \emph{Chaturaji}. Like the green cluster, this group of games is not characterised by a single concept. The key aspects that group these games together, seem to be the combination of a large number of pieces for each side, as well as complex movement rules for each individual piece.
    \item The blue cluster contains all other games which do not fall into the previous clusters.
\end{itemize}

 From this, we can first see that the blue cluster is considerably larger than the others, making up more than half the total number of games. While this cluster could probably be split up further into sub-clusters, the separation is not as clear as for the clusters identified. Admittedly, the separation between the blue and red clusters is also not as distinct as the others, but is clear enough that we felt it worth mentioning.
 With the exception of the orange cluster, which is uniquely defined by the existence of the "Mancala Board" concept, there is no single concept that is responsible for any one cluster. Each cluster is instead defined by a combination of multiple concept values, making previous attempts to categorise games based on singular properties incapable of creating such a distinction.
 Based on these findings, it is clear that the recorded concept values provide significant information about a game's mechanics and properties, and are likely to be an effective basis for measuring the distance between two games.

\section{Game Distance}
\label{Game Distance}

When it comes to calculating the distances between games, each game is represented as a vector of 499 normalised concept values. Comparing two games can therefore be done using a variety of different vector distance/similarity measures. This includes Euclidean distance, Manhattan distance, Cosine similarity, Jaccard index, Jenson-Shannon divergence, among many other options. 

Additional pre-processing can also be applied to adjust the importance of each concept. For example, Inverse Document Frequency (IDF) could be applied to all binary concepts, increasing the weights for concepts that occur in fewer games while decreasing the weights for those that occur in many. 
Each concept category could also be adjusted as a whole. For example, each concept could be scaled relative to the size of its category, resulting in each category carrying equal collective weighting. 
Some categories could even be excluded completely, as the importance of each category will likely vary based on the intended application. For example the "Visual" category of concepts has no bearing on actual gameplay, and would provide no benefit for the application of training a general game-playing agent.

Unfortunately, due the lack of any concrete benchmarks for measuring game distance, the effectiveness of these different measures and weight adjustments cannot be objectively evaluated without a specific application in mind. Due to the open-ended nature of this exploratory study, we decided to only compare the Cosine similarity and Euclidean distance measures, two of the most popular vector distance measures, without any additional pre-processing or weight adjustments. We encourage other researchers who wish to use this dataset for their own work, to experiment with these different distance measure approaches and identify what works best for their desired application.

The normalised Cosine distance between two games $G_1$ and $G_2$, with concept vectors denoted by $\vec{\bf c}_{1}$ and $\vec{\bf c}_{2}$ respectively, is given by \refequation{Eq:CosineDistance}.

\begin{equation} \label{Eq:CosineDistance}
Cosine Distance(G_{1}, G_{2}) = \frac{1}{2} \left(1-\frac{\vec{\bf c}_{1} \cdot \vec{\bf c}_{2}}  {\|\vec{\bf c}_{1}\| \|\vec{\bf c}_{2}\|}\right)
\end{equation}

The normalised Euclidean distance between two games, using the same terms in the previous equation and $n$ denoting the total number of concepts, is given by \refequation{Eq:EuclideanDistance}.

\begin{equation} \label{Eq:EuclideanDistance}
Euclidean Distance(G_{1}, G_{2}) = \frac{\|\vec{\bf c}_{1} - \vec{\bf c}_{2}\|} {\sqrt{n}}
\end{equation}

Both of these distance measures are normalised within the zero to one range, with one representing maximal difference and zero representing maximal similarity.

\section{Results}
\label{Results}

Both Euclidean and Cosine distances were calculated between each possible pair of our 1059 games, giving a total of 560,211 unique game pairs for each distance measure.

\begin{figure}[t]
\centerline{\includegraphics[width=0.7\linewidth]{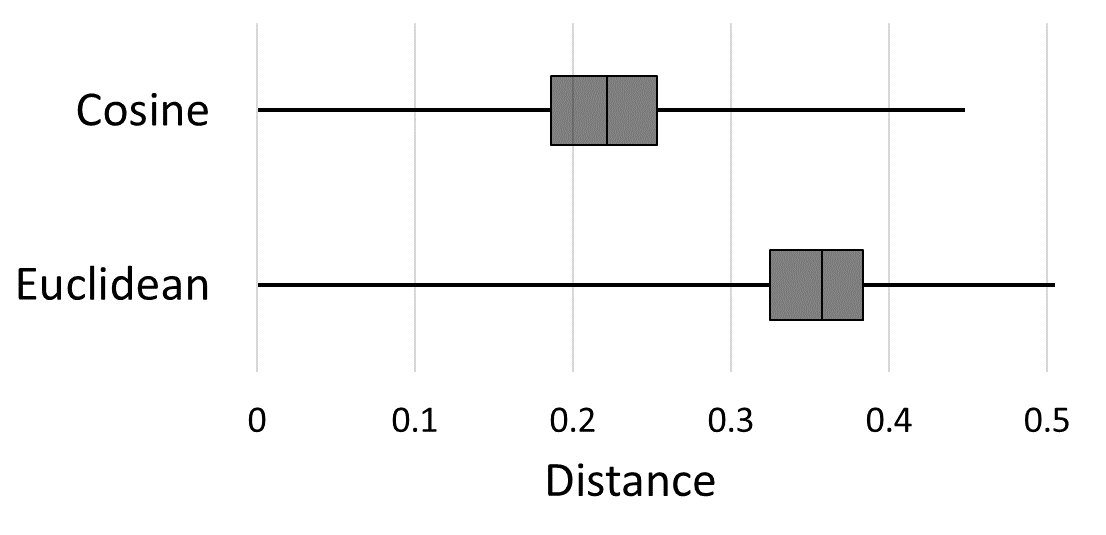}}
\caption{Box plot for Cosine and Euclidean distances across all game pairs.}
\label{boxplot}
\end{figure}

\reffigure{boxplot} presents a box plot visualisation of each distance measure across all game pairs.
From this, we can see that the Euclidean distance is typically larger than the Cosine distance. The inter-quartile range for Cosine distance is situated almost exactly equally between its minimum and maximum values, while the inter-quartile range for Euclidean distance is skewed much closer to the maximum value. The difference between the median values of each distance measure (0.1358) is also much larger than the difference between their maximum values (0.0572).  

\begin{figure}[t]
\centerline{\includegraphics[width=0.75\linewidth]{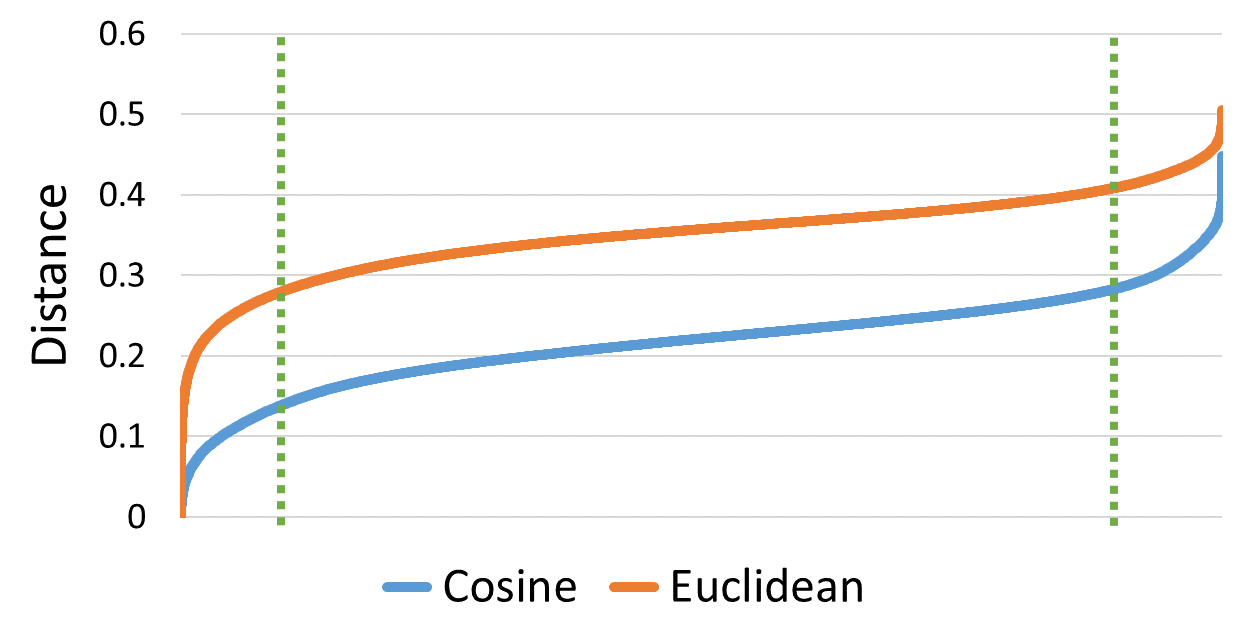}}
\caption{Cosine and Euclidean distance trends, ordered by size along the x-axis.}
\label{trendChart}
\end{figure}

\reffigure{trendChart} provides further details on this observation, showing the general trend for each distance measure across all game pairs when ordered from smallest to largest along the x-axis. Looking at the 10\% - 90\% interpercentile range, represented by the area between the two green dashed lines, shows that the gradient of each distance measure is approximately equal.
The most significant difference between the trends of these two distance measures is instead located at the extremities. While both distance measures begin at zero, the Euclidean distance initially increases far more rapidly than the Cosine distance. The opposite is true at the upper percentile end, with the Cosine distance taking a sharp increase to raise itself closer to the Euclidean distance.

\begin{figure}[t]
\centerline{\includegraphics[width=0.75\linewidth]{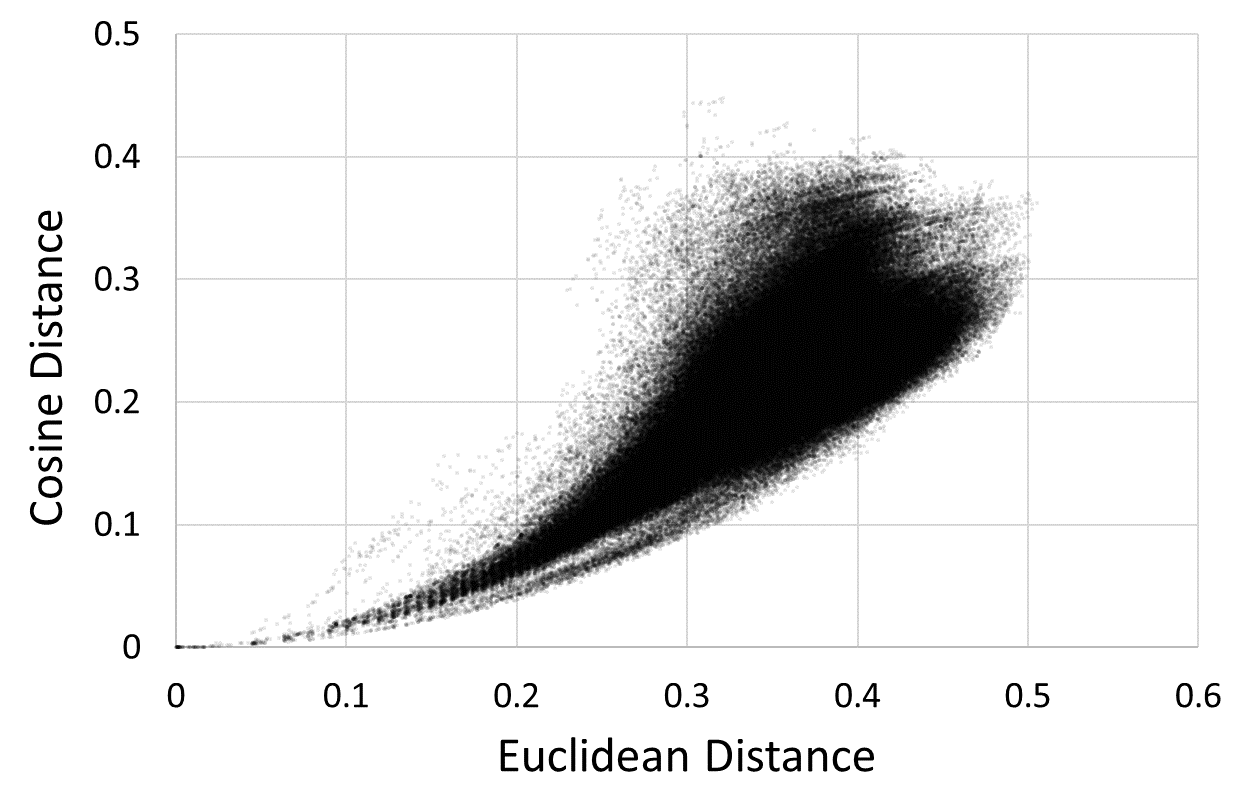}}
\caption{The Cosine and Euclidean distance for all 560,211 game pairs.}
\label{allGamePairs}
\end{figure}

\reffigure{allGamePairs} visualises the differences between the Cosine and Euclidean distances for each individual game pair. From this we can see that there is a strong positive relationship between both distance measures, with a Pearson correlation coefficient of 0.7574. 
The overall upward curve of the points also reiterates our prior observation, that the rate at which the Euclidean distance increases is initially much higher, but then gradually falls to be more in line with that of the Cosine distance.

\subsection{Discussion}

While these statistical results provide a broad overview of how these distance measures compare across all games, we also explore how they differ with regard to specific game pairs.
The game pair with the greatest difference in their Cosine and Euclidean distances (with a higher Cosine distance) was between \emph{Magic Square} and \emph{Rock-Paper-Scissors}, with a Cosine distance of 0.4438 and a Euclidean distance of 0.3036. Both of these games are relatively simple and share very little in common, with \emph{Magic Square} being a logic puzzle and \emph{Rock-Paper-Scissors} being a two-player game with simultaneous moves.
From the alternative perspective, the game pair with the greatest difference in their Euclidean and Cosine distances (with a higher Euclidean distance) was between \emph{Tenjiku Shogi} and \emph{Chex}, with a Cosine distance of 0.1429 and a Euclidean distance of 0.3792.
Both of these games are very complex, featuring large boards along with many different pieces and rules.

Based on these two contrasting game pair examples, it initially seems that Cosine distance is greatest between simpler games with relatively few high-value concepts, while Euclidean distance is greatest between more complex games with a larger number of high-value concepts. 

To dive deeper into how each distance measure compares across multiple game pairs, we looked at which game pairs received the largest values from each measure. One immediate observation was that the majority of the largest Cosine and Euclidean distances were between a logic puzzle, such as \emph{Sudoku}, \emph{Kakuro} or \emph{Hoshi}, and non-puzzle game. This significant logic puzzle presence makes intuitive sense, as they are a unique type of game that would likely produce very distinct concept values.

For the Cosine distance measure, all of the 20 largest game pair distances involved either \emph{Rock-Paper-Scissors}, \emph{Morra} or \emph{Aksadyuta}. All three of these games are very simple in terms of their rules, and are essentially purely random in terms of their outcome. These games contain no boards or pieces (at least in the traditional sense), and typically last for only a few moves. It would therefore make sense for these games to be highly distant from most other games, and further backs up our theory that the Cosine distance measure gives the greatest distance values to game pairs that include a simpler game with less common concepts.

For the Euclidean distance measure, all of the 20 largest game pair distances involved either \emph{Beirut Chess}, \emph{Ultimate Chess} or \emph{Chex}. These games are all Chess variants with some unique twist on their rules. 
All three of these games would probably be considered more complex than the majority of other games in our dataset. This likely results in a substantial number of large value concepts for each game, again supporting the idea that the greatest Euclidean distances are typically given to game pairs that involve at least one high complexity game.

Based on these further comparisons, it appears that neither distance measure is inherently better than the other. It instead seems likely that each approach, as well as the other suggested measures that we did not explore deeper, has its own strengths, weaknesses and biases.
The choice of which distance measure is most suitable is a highly subjective decision, and would depend on the intended application. We therefore reiterate our previous statement that multiple approaches should be tested and evaluated for each specific use-case, rather than attempting to develop a single correct measure of game distance.


\section{Conclusion}
\label{Conclusion}

In this paper, we have investigated the use of general board game concepts to measure the distance between pairs of games. Based on the same original concept dataset, two different measurements were proposed based on Cosine and Euclidean distance. Our results highlight the differences between these approaches. Cosine distance tended to give its highest values to pairs of relatively simple games with very few shared concepts. Euclidean distance on the other hand, appeared to include larger and more complex games in its highest value pairs, where any similarities between the games were outweighed by their differences. While we are unable to conclude which distance measure might be better or worse for any specific application, the contrasting outputs between these two distance measures illustrates the importance of experimenting with multiple distance measurement approaches.

Possible future work could involve a more complete analysis and summary of a larger range of distance measurements, as well as the effect that different pre-processing and weight adjustment techniques has on their outputs. Additional concepts could also be added to fill knowledge gaps within the existing dataset. One addition could be the inclusion of playout concepts based on alternative game traces, such as those produced from different game-playing agents or human players. Rather than adding more concepts, a more nuanced and critical look at the existing corpus may instead lead to the removal or weight reduction of certain items. It may not make sense to treat meaningful concepts, such as whether a game involves hidden or stochastic information, with as much importance as overly niche concepts, such as whether the game includes pieces that move backwards to the left. However, such alterations to the concept dataset are likely to be application specific, as adding or removing concepts for one purpose may inadvertently affect another. 

\section*{Acknowledgements}

This research is funded by the European Research Council as part of the Digital Ludeme Project (ERC Consolidator Grant \#771292) led by Cameron Browne at Maastricht University's Department of Advanced Computing Sciences.

\bibliographystyle{IEEEtran}
\bibliography{dlp-biblio-1}

\end{document}